\documentclass[conference]{IEEEtran}
\IEEEoverridecommandlockouts
\usepackage{cite}
\usepackage{amsmath,amssymb,amsfonts}
\usepackage{algorithmic}
\usepackage{graphicx}
\usepackage{textcomp}

\usepackage{hyperref}

\usepackage{xcolor}
\usepackage[caption=false]{subfig}
\def\BibTeX{{\rm B\kern-.05em{\sc i\kern-.025em b}\kern-.08em
    T\kern-.1667em\lower.7ex\hbox{E}\kern-.125emX}}
\begin{document}

\title{KIT Bus: A Shuttle Model for CARLA Simulator}

\makeatletter
\newcommand{\linebreakand}{%
  \end{@IEEEauthorhalign}
  \hfill\mbox{}\par
  \mbox{}\hfill\begin{@IEEEauthorhalign}
}

\makeatother

\author{
  \IEEEauthorblockN{1\textsuperscript{st} Yusheng Xiang}
  \IEEEauthorblockA{\textit{Institute of Vehicle System Technology} \\
    \textit{Karlsruhe Institute of Technology}\\
    Karlsruhe, Germany \\
    yusheng.xiang@partner.kit.edu}
  \and
  \IEEEauthorblockN{2\textsuperscript{nd} Shuo Wang }
  \IEEEauthorblockA{\textit{Institute of Vehicle System Technology} \\
    \textit{Karlsruhe Institute of Technology}\\
    Karlsruhe, Germany \\
    shuo.wang@kit.edu}
  \and
  \IEEEauthorblockN{3\textsuperscript{rd} Tianqing Su}
      \IEEEauthorblockA{\textit{Guanghua School of Management} \\
    \textit{Peking University}\\
    Beijing, China \\
    tianqing.su@hotmail.com}
  \linebreakand 
  \IEEEauthorblockN{4\textsuperscript{th} Jun Li}
  \IEEEauthorblockA{\textit{Center of AI} \\
    \textit{University of Technology Sydney}\\
     Ultimo, Australia \\
    jun.li@uts.edu.au}
  \and
  \IEEEauthorblockN{5\textsuperscript{th} Samuel S. Mao}
  \IEEEauthorblockA{\textit{Department of Mechanical Engineering} \\
    \textit{University of California, Berkeley}\\
    Berkeley, USA \\
    ssmao@berkeley.edu}
  \and
  \IEEEauthorblockN{6\textsuperscript{th} Marcus Geimer}
  \IEEEauthorblockA{\textit{Institute of Vehicle System Technology} \\
    \textit{Karlsruhe Institute of Technology}\\
    Karlsruhe, Germany \\
    marcus.geimer@kit.edu}
}
\maketitle

\begin{abstract}

With the continuous development of science and technology, self-driving vehicles will surely change the nature of transportation and realize the automotive industry's transformation in the future. Compared with self-driving cars, self-driving buses are more efficient in carrying passengers and more environmentally friendly in terms of energy consumption. Therefore, it is speculated that in the future, self-driving buses will become more and more important. As a simulator for autonomous driving research, the CARLA simulator can help people accumulate experience in autonomous driving technology faster and safer. However, a shortcoming is that there is no modern bus model in the CARLA simulator. Consequently, people cannot simulate autonomous driving on buses or the scenarios interacting with buses. Therefore, we built a bus model in 3ds Max software and imported it into the CARLA to fill this gap. Our model, namely KIT bus, is proven to work in the CARLA by testing it with the autopilot simulation. The video demo is shown on our \href{https://www.youtube.com/watch?v=UwLRW8gDJGw}{Youtube}.


\end{abstract}



\section{Introduction}

Autonomous driving buses are an important part of the future metropolis. Compared with a passenger car, a bus has a larger size, which, on the one hand, makes it has the potential to reduce the operating cost per passenger and increase the traffic efficiency \cite{Dong&DiScenna&Guerra.Tupodb.2019}. On the other hand, it is more difficult to recognize a complete view of the buses' surrounding environment. Consequently, more sensors should be equipped on an autonomous bus in order to cover all important areas, avoid blind spots, and react in complicated scenarios. The confusion of sensors like cameras, radar, LIDAR is therefore necessary. However, current studies about autonomous shuttle depend mostly on the limited experiments in reality, such as \cite{Montes&Salinas.AEPfABD.2017, Millonig&Frohlich.WABMaMNBtGit4AsoPTPN.09232018}. Consequently, a comprehensive analysis of the sensors arrangement for an autonomous-standard bus does not exist. Obviously, due to insufficient data, regulations, policies, and legal issues are challenging to be deeply explored. To date, gathering the data and validating novel ideas in simulation environments are becoming more and more popular due to the high fidelity of the autonomous driving simulator, e.g., the CARLA simulator \cite{Dosovitskiy&Ros.CAOUDS.2017,Dworak&Ciepiela.PoLoddlaboagpcdfCs.20198262019829}. However, the current official CARLA does not provide a bus model that can offer the opportunity to research self-driving buses. Thus, in this study, we created a large-size bus model for the CARLA simulator to be used as a cornerstone for further studying the autonomous driving bus. As shown in Fig. \ref{thesisbusmodel}, our model enables the high fidelity simulation, including physical engine, e.g., collision and occlusion, in scenarios with buses. 

\begin{figure}[tbp]
\centerline{\includegraphics[width=3.5in]{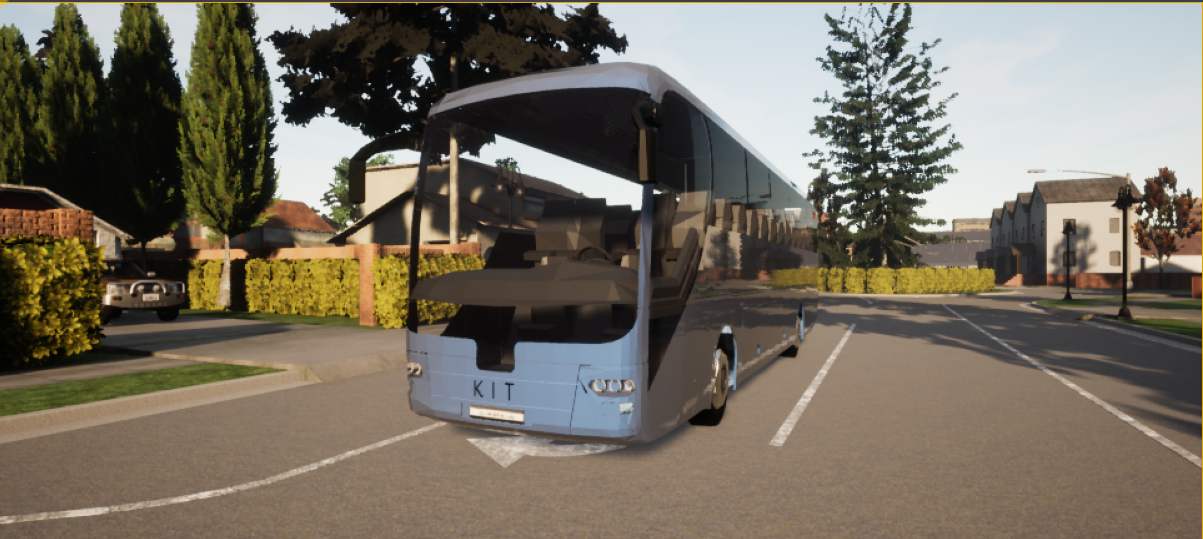}}
\caption{The KIT Bus in the CARLA simulator}
\label{thesisbusmodel}
\end{figure}

\section{Background}

Although the shuttle bus transport capacities currently in use are low, i.e., usually only 15 people, the use of these transport units in large urban agglomerations is beneficial for society \cite{Iclodean&Cordos&Varga.ASBfPTAR.2020}. In contrast to current wisdom, shared fleets may not be the most efficient alternative \cite{Bosch&Becker.Cbaoams.2018}. Based on the research from Leich, autonomous shared taxis cannot replace buses since their simulation suggests for all evaluated scenarios higher operating costs and only slight travel time savings in comparison to conventional buses \cite{Leich&Bischoff.SastrbAss.2018}. In 2017, a survey among 200 passengers of autonomous mini-buses in real operation in the city of Trikala showed that they are well accepted, and there are no major concerns as regards safety and security \cite{Portouli&Karaseitanidis.PatamboirciaHc.2017}. Also, the platooning technology endows the buses to handle the scenarios where complex interactions occur \cite{Lam&Katupitiya.Macoapoab.20136232013626}. 

Today's self-driving buses use various cutting-edge hardware and software technologies to complete their driving. A typical autonomous driving system will go through three stages to perform its driving, called perception \cite{Gruyer&Magnier.PipamCsfada.2017}, understanding \cite{Chen&Jian.Adccasu.2019}, and control. In the perception  phase, cameras and various sensors are used to observe objects around the vehicle, such as other vehicles, people, bicycles, and animals. The sensor perception technologies currently studied include visual perception, laser perception, microwave perception, etc. In the understanding stage, various artificial intelligence algorithms are used to process information from sensors. In the control stage, the autonomous driving system will process all the information that the perception system can extract. Currently, a popular way to tackle the lack of data is to use a game engine or a high-fidelity computer graphics model to create driving scenes. At present, the simulation system can use virtual lidars and cameras to scan a street section to simulate traffic conditions that are highly similar to the real situation. At the same time, advanced view synthesis technology can generate scene images of any perspective based on existing images, which looks as real as direct shooting \cite{LiW.PanC.W.ZhangR.RenJ.P.MaY.X.FangJ.YanF.L.GengQ.C.HuangX.Y.Gon.2AAadsudda42.2019}. More survey papers about self-driving buses can be found \cite{Azad&Hoseinzadeh.FABALRaFRD.2019, Ainsalu&Arffman.SotAoAB.2018}.

\section{Model Building}
In this section, we demonstrate the details of the white model building, as shown in Fig. \ref{fig:process of the model building}. 

As the first step, we found a drawing size of the reference bus model in the Internet, including the actual length, width, and height of the bus, as shown in Fig. \ref{fig:process of the model building} (a). In order to build a three-dimensional model of the bus, we cut the size map of the bus model into four bitmaps and created four bus model reference planes in 3ds Max. The size of each plane corresponds to the length, width, and height of the bus model and map the four bitmaps on the planes. In this fashion, the size of the four planes conforms to the size of the created bus model equal to the real bus size. Secondly, we build the bus model based on it. We use the line tool to draw the outline of the bus body based on the front view and then extrude half of the bus body with reference to the left view or the right view, as shown in Fig. \ref{fig:process of the model building} (b). 

\begin{figure}[!t]
        \newcommand{\w}{0.45}
        \centering 
        \subfloat[3D dimensional drawing of a bus model from MAN AG for reference. The dimension of the bus is $ 12m \times 2.55m \times 3.81m $ in terms of length, width, and height, respectively. The capacity of the selected bus is designed for about 45 passengers. ]{\includegraphics[width=\w\textwidth]{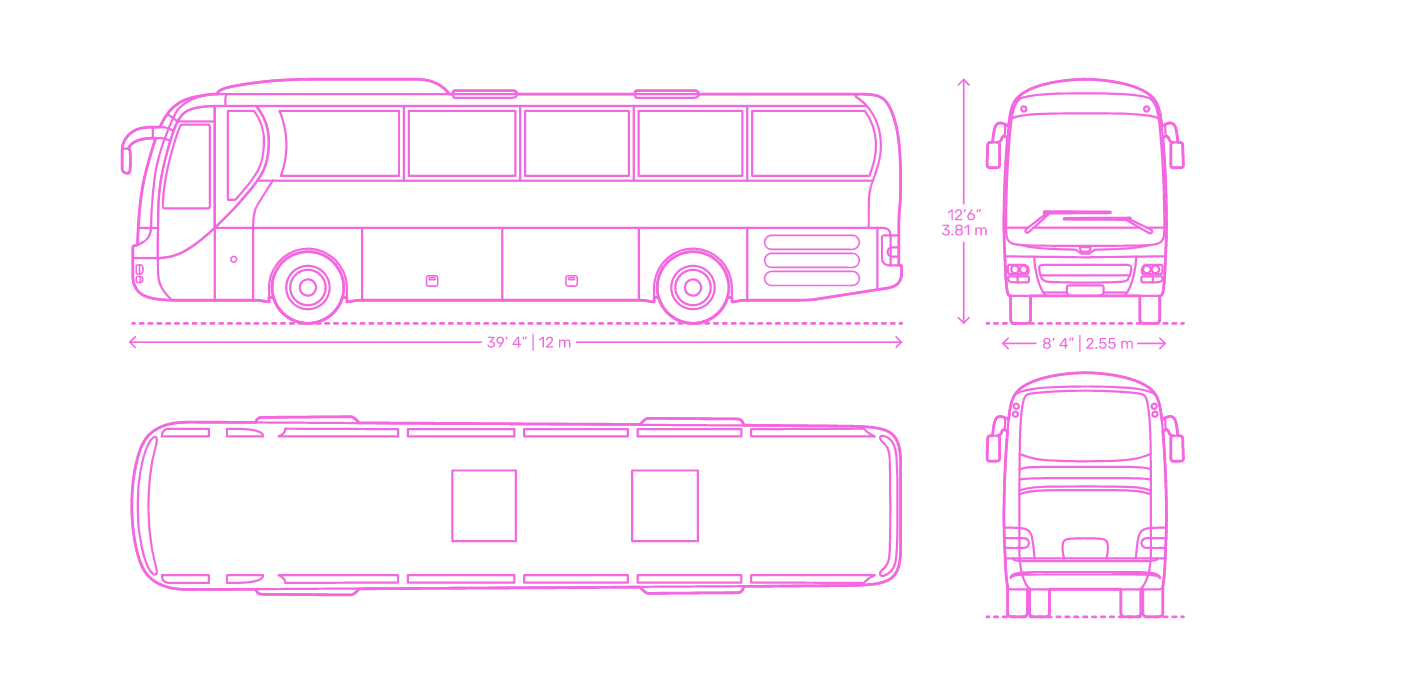}}
            \hfil 
        \subfloat[Drawing outline and extrusion]{
            \includegraphics[width=\w\textwidth]{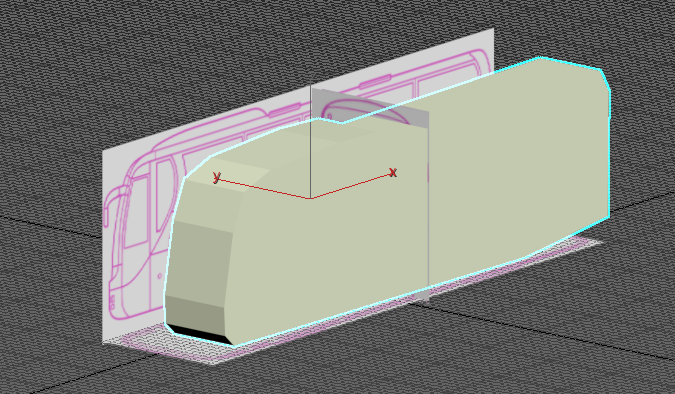}} 
            \hfil
        \subfloat[Bus model without air conditioner]{
            \includegraphics[width=\w\textwidth]{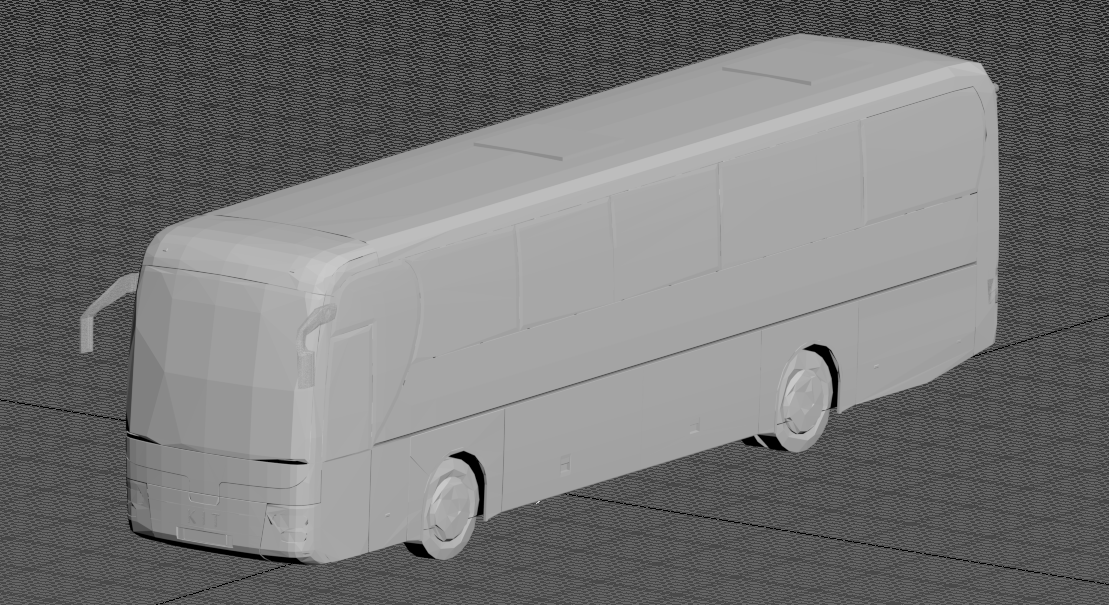}} 
            \hfil
        \caption{The process of building up the KIT bus} 
        \label{fig:process of the model building}
\end{figure}

After that, we divide the extruded body into small segments by creating points, lines, and connecting points. Concretely, we use the function of moving points, lines, and surfaces to make the divided lines as closely as possible according to the contours of the three views in different directions. By fitting the contours of the three views of the bus model and continuously making detailed adjustments to ensure that the surface of the bus mode is as smooth as possible, we improve the similarity between our bus model and the real bus as much as possible. Through continuous refinement, we obtain the completed bus body model. 

Lastly, we build the rest of the bus model parts in a similar way to bus body modeling. To sum up, the bus model consists of six parts, e.g., body, wheels, interior, details, glass, and license plate. Wheels include tires and wheel hubs. The interior includes the seats, the steering wheel, and the bottom of the bus. Details include lights, rearview mirror, and logo. Glass includes light glasses, windows. Fig. \ref{fig:process of the model building} (c) is the frame of the final bus model.


\section{Setup our model}

This section depicts the process from downloading our model (see Appendix) until running the model in the CARLA simulator. 

First, download our model from Google drive and import it into the static model file in CarlaUE4. After finishing the import process, the skeletal mesh will appear along with two new files, called Bus\_Skeleton and Bus\_PhysicsAssets, as shown in Fig. \ref{import} (a). 

\begin{figure*}[!t]
        \newcommand{\w}{0.45}
        \centering 
        \subfloat[Import bus model into CarlaUE4]{
            \includegraphics[width=\w\textwidth]{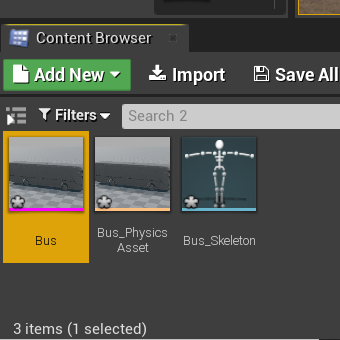}} 
            \hfil
        \subfloat[Collider of bus body]{
            \includegraphics[width=\w\textwidth]{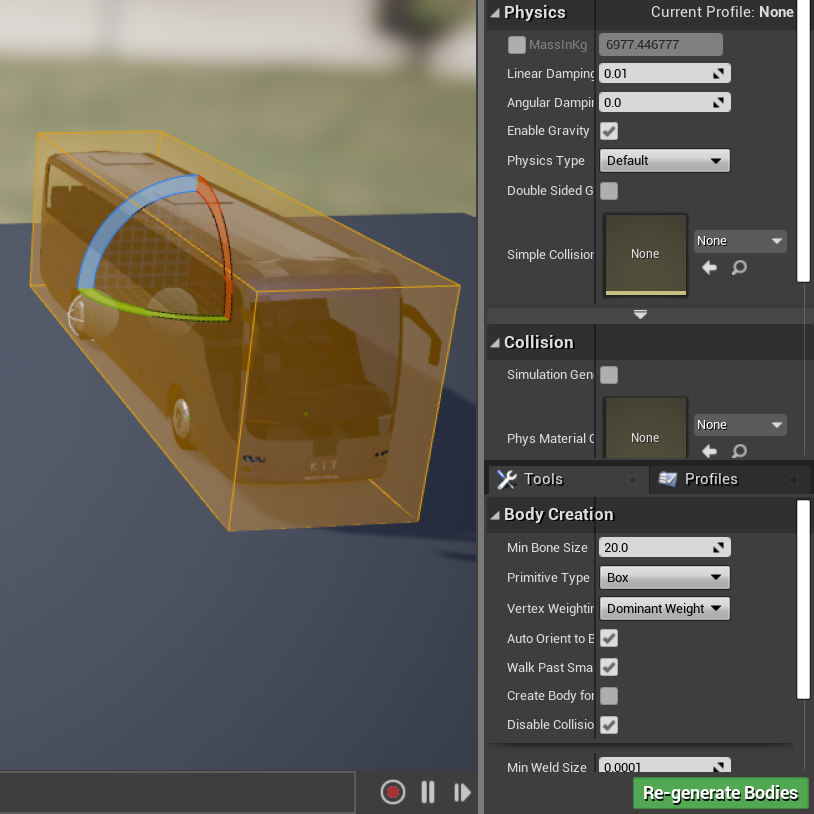}}
            \hfil 
        \subfloat[Collider of bus wheels]{
            \includegraphics[width=\w\textwidth]{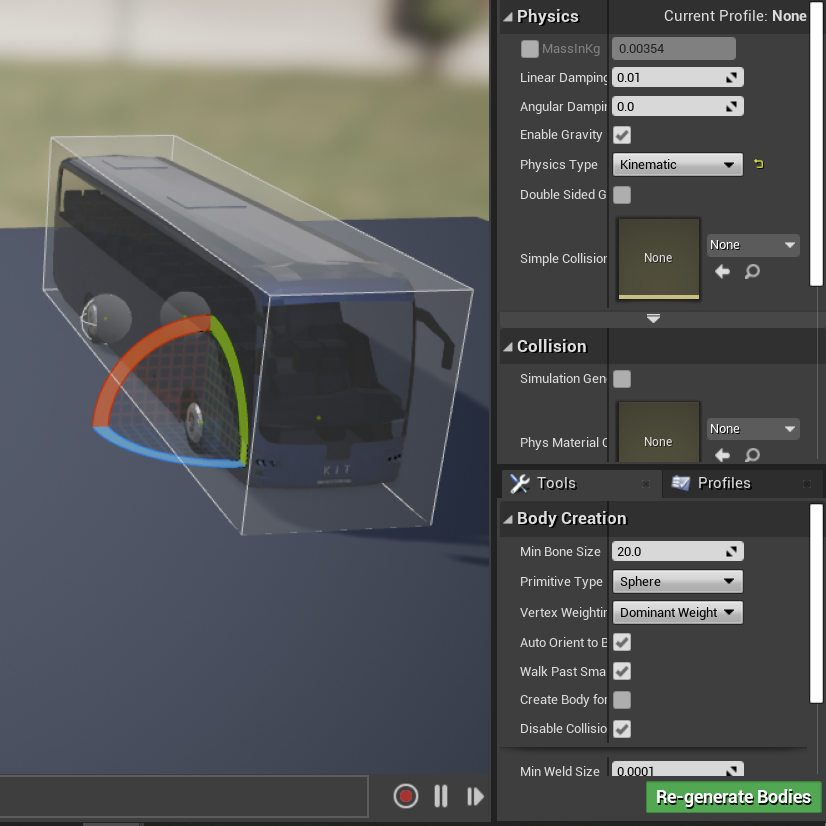}} 
            \hfil
        \subfloat[Create the animation blueprint]{
            \includegraphics[width=\w\textwidth]{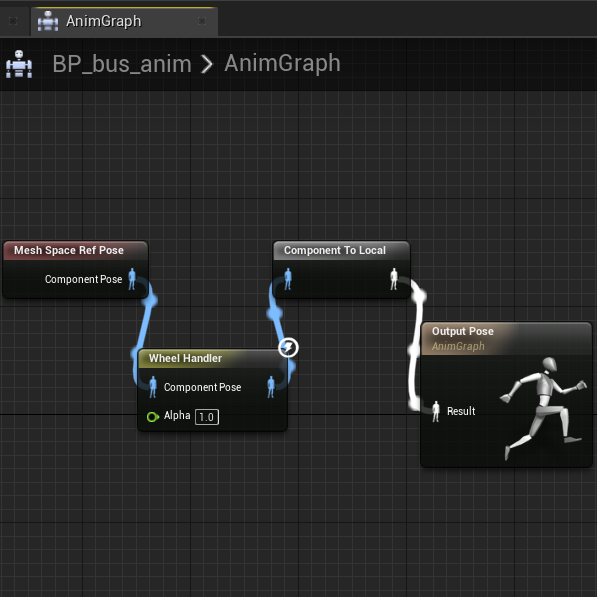}} 
            \hfil
        \caption{Import and link process} \label{import}
\end{figure*}


Secondly, set up collisions in Bus\_PhysicsAsset. Since the body of the bus model is a rectangular parallelepiped shape, so we choose the primitive type of the Vehicle\_Base as the box, and regenerate the body, as shown in Fig. \ref{import} (b). Then we set the bus model wheels colliders. Because the shape of the bus model wheel is round, we choose the primitive type of the four wheels as sphere and change the physics type as kinematic, and then again regenerate the body, as shown in Fig. \ref{import} (c). Thirdly, create the bus model animation blueprint. In order for the bus model to simulate real bus driving animation in CarlaUE4, we must create an animation blueprint. As the driving characteristics of the bus model depends on its wheels, we create an animation blueprint for the wheel handler. As shown in Fig. \ref{import} (d), it is the process of creating a blueprint for the bus model animation. It includes four animation blueprint nodes, namely mesh space ref pose, wheel handler, component to local, and output pose. The connection of the mesh space ref pose node and the wheel handler can link the animation of the bus model wheels with the skeletal mesh of the entire model. 


\begin{figure*}[!t]
        \newcommand{\w}{0.45}
        \centering 
        \subfloat[Create front wheel blueprint]{
            \includegraphics[width=\w\textwidth]{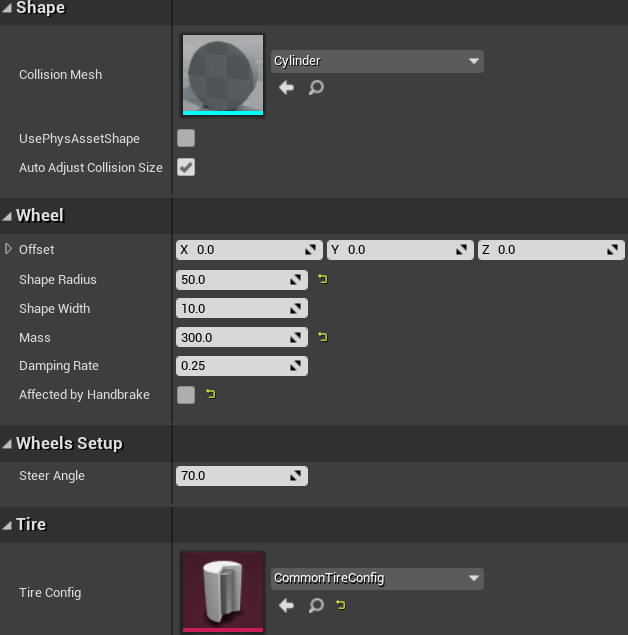}} 
            \hfil
        \subfloat[Create rear wheel blueprint]{
            \includegraphics[width=\w\textwidth]{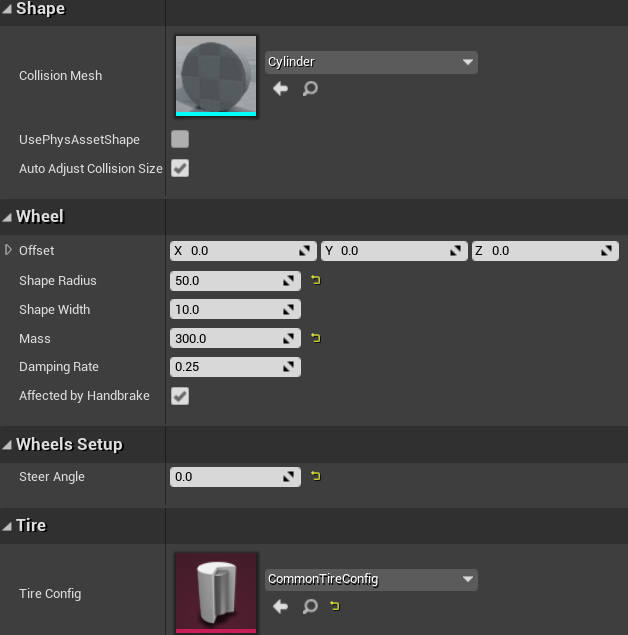}}
            \hfil 
        \caption{Prepare the blueprint of wheels} \label{wheels}
\end{figure*}

Next,  two wheel-blueprints shall be created. In most cases, the bus model shall have at least two types of wheels, steering wheels and driving wheels.  In addition, there may be front and rear wheels of different sizes. In this case, we need to control the settings of different radius, mass, width, handbrake effect, suspension, and many other attributes to get the bus control we need. Therefore, we create two wheel-blueprints, the front wheel blueprint, and the rear wheel blueprint. The front wheel is not affected by the handbrake whilst it can be steered, as shown in Fig. \ref{wheels} (a), and the rear wheel is affected by the handbrake but cannot be steered, as shown in Fig. \ref{wheels} (b). The tire config in the two figures is used to set the value of the loaded tire properties, so we need to import the common tire config in our case.



Obviously, in addition to creating a blueprint for the wheels, we need to create a blueprint for the bus model as a whole. The purpose is to match the previous skeleton mesh with the animation blueprint so that the animation blueprint can control the skeleton mesh,  as shown in Fig. \ref{busblueprint}.

\begin{figure}[htbp]
\centerline{\includegraphics[width=8cm]{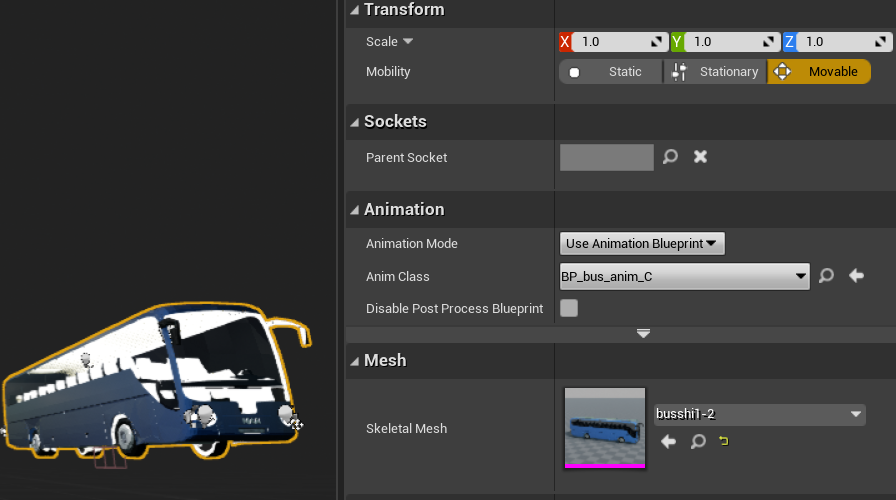}}
\caption{Blueprint for the bus model}
\label{busblueprint}
\end{figure}

After creating the blueprint of the bus model, we need to match the four wheels together in this blueprint in order to realize that the different wheel positions of the bus can drive according to the parameters we set, as shown in Fig. \ref{wheelsetup}.

\begin{figure}[htbp]
\centerline{\includegraphics[width=8cm]{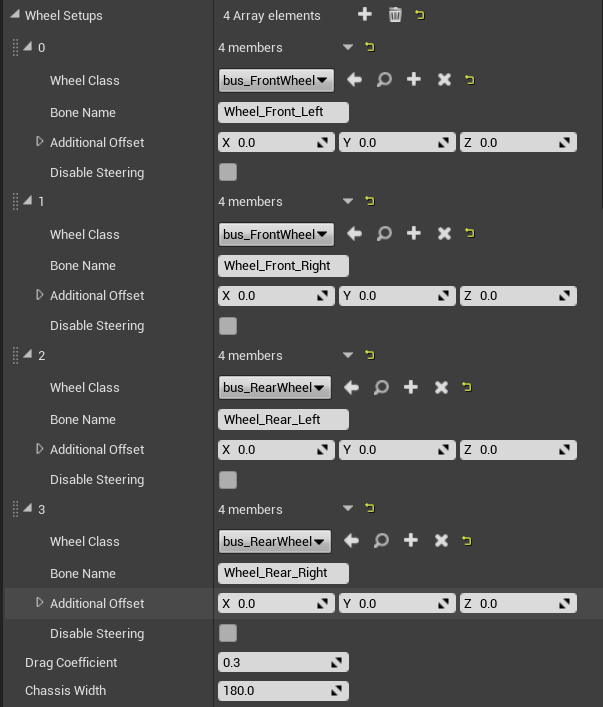}}
\caption{ Wheels set up}
\label{wheelsetup}
\end{figure}

Finally, we edit the material of the bus model and add functions according to the needs, such as adding different types of bus lights and so on. This step is extremely time-consuming. Our bus model is divided into thirteen parts that can be edited with materials: the metal body, bumper, rearview mirror cover, windows, tires, wheels, trunk handles, taillights, headlight glass covers, front lights, interior trim, rearview mirror, license plate. Concretely, we use the material editor for material rendering in CarlaUE4. The main method of operation is to splice each characteristic node with the result node of material through a line. By observing the display 
result of the material texture in the preview of the material ball, we continuously modify and adjust the parameters until the material texture displays bus model very close to the real-world bus material appearance. As shown in Fig. \ref{uebusmodell}, the final bus model with proper materials setting is created. Notice that this step makes our bus model look similar to the real bus and thus can be easier recognized by the visual-based artificial intelligence model. However, for other tasks, such as the appropriate arrangement of sensors, it is not so vital.  

\begin{figure}[htbp]
\centerline{\includegraphics[width=8cm]{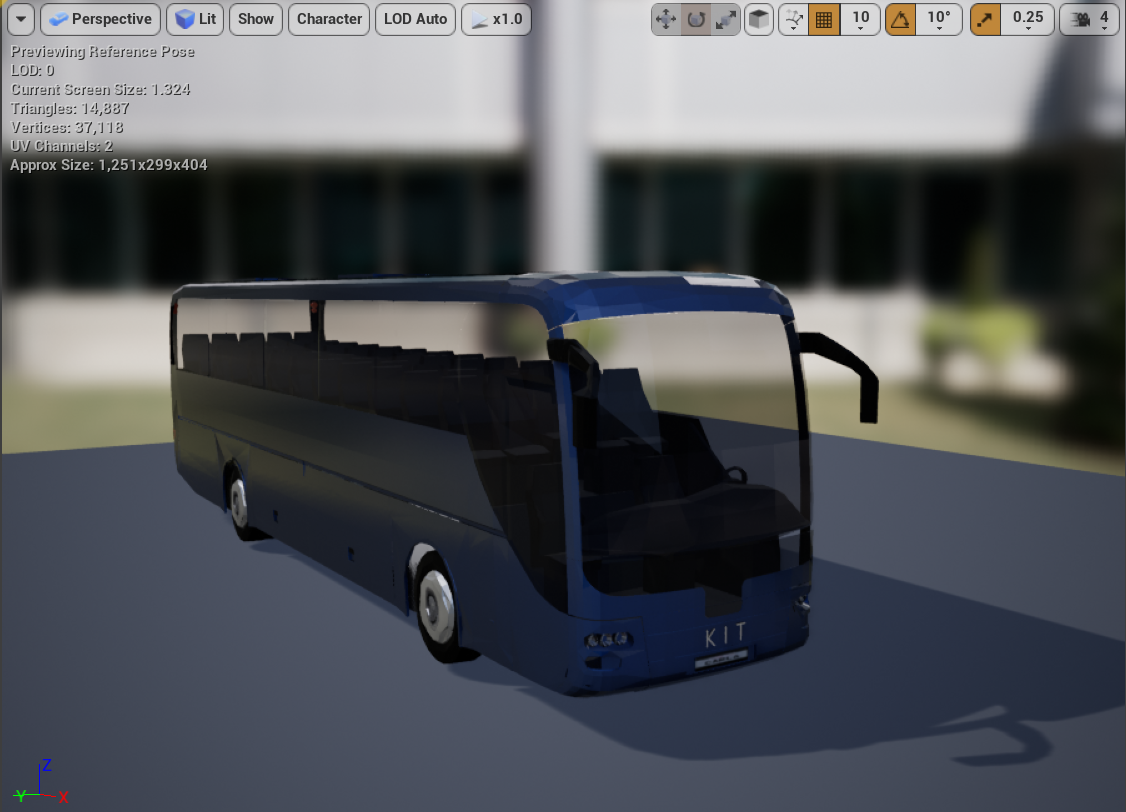}}
\caption{ Bus model}
\label{uebusmodell}
\end{figure}

Finally, we have completed the material editing of the bus in CarlaUE4 and the creation and editing of various blueprints for animation. We need to put the edited bus model into the vehicle factory of the CARLA simulator before we can control it, as shown in Fig. \ref{vehiclefactory}. 

\begin{figure}[htbp]
\centerline{\includegraphics[width=8cm]{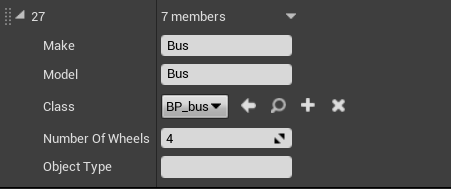}}
\caption{Vehicle Factory}
\label{vehiclefactory}
\end{figure}

\section{Validate the model in CARLA simulator}

After successfully importing the model in the CARLA simulator, do the following steps to see whether the model works. Real-world bus driving functions mainly include forwarding, braking to decelerate, steering, reversing, and gear shifting. With these functions, we can control the bus model by using python scripts and realize different buttons to control different motion functions through KeyboardControl. Although we tested our model with autopilot, we give the process to run the model with the manual operation since it gives the readers more freedom to test the model.

As the first step, run the python code of the manual control, which is in the PythonAPI folder and can put the bus model into an environment that simulates the real world, as shown in Fig. \ref{pygamewindow}.

\begin{figure}[htbp]
\centerline{\includegraphics[width=8cm]{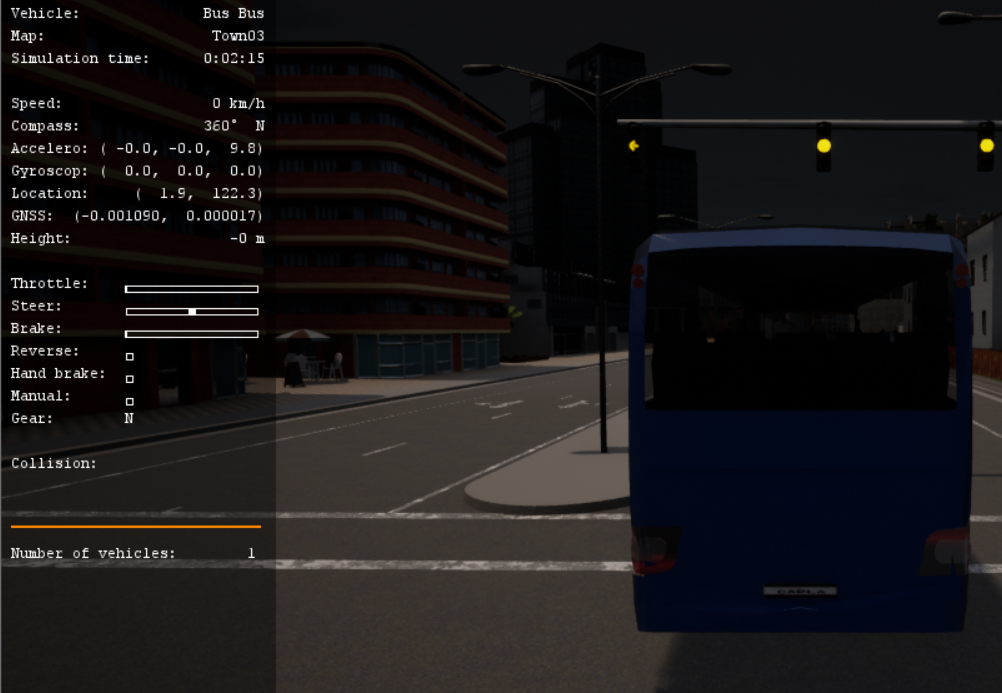}}
\caption{Import the model into the CARLA simulator scene}
\label{pygamewindow}
\end{figure}

Now the model can be controlled by the keyboard. With the W key to control the bus to move forward, the information bar on the left side of the screen displays the throttle change during the forwarding process. In contrast, the S key controls the braking process of the bus. Similar to the driving command, the dashboard also displays the braking force during the braking process. What's more, the A key and the D key control the bus to turn left and right, respectively. Here the cursor position demonstrates the angle and direction of the steering. Apparently, when the cursor is in the middle, it means to go straight forward. The Q key controls the bus to reverse. The space key controls the hand brake. Once turn on the hand brake, the vehicle is forcibly stopped. The M key switches the automatically shifting or manual gear shifting.  With the ``," and ``." keys, we make the model upshift and downshift, respectively.


\section{Conclusion}

We have successfully built a bus model for the CARLA simulator that can be used for autonomous driving simulation. In this process, we are mainly divided into three main steps. Firstly, we use the three-view modeling method to build the bus model with 3ds Max software. We use the line tool to draw the outline of the bus parts based on the front view in the three views and extrude half of the bus parts regarding the left view or the right view. Then we create points, lines, and use the function of moving points, lines, and surfaces to make the divided lines according to the contours of the three views in different directions. By fitting the contours of the three views of the bus model, and continuously making detailed adjustments, we ensure that the surface of the bus body is as smooth as possible. After we get the bus geometry model, we edit the bus model in CarlaUE4. It includes the main five processes, bus model material editing, setting the bus model colliders, creating the animation blueprint, creating blueprints for wheels and bus model, and adding additional light functions. Finally, we verified that the built model can be controlled to drive both manually and autonomously in the CARLA simulator.

This study only considered the autonomous driving simulation of one bus model alone and did not conduct in-depth explorations on the driving of multiple vehicles that simulate the real traffic. However, we envision the research in the next studies. 

\section*{Appendix}

We can share the model we built. Check the most updated information about the model on the Elephant Tech LLC website~\footnote{\href{http://www.elephant-tech.cn/}{http://www.elephant-tech.cn/}}.  Please contact us with your self introduction and a brief description (one page) of your research plan. Commercial use and redistribution without permission is prohibited. Only Yusheng Xiang will reply to the emails related to this paper. Thank you.





\bibliography{literature.bib}{}

\begin{thebibliography}{10}
\providecommand{\url}[1]{#1}
\csname url@samestyle\endcsname
\providecommand{\newblock}{\relax}
\providecommand{\bibinfo}[2]{#2}
\providecommand{\BIBentrySTDinterwordspacing}{\spaceskip=0pt\relax}
\providecommand{\BIBentryALTinterwordstretchfactor}{4}
\providecommand{\BIBentryALTinterwordspacing}{\spaceskip=\fontdimen2\font plus
\BIBentryALTinterwordstretchfactor\fontdimen3\font minus
  \fontdimen4\font\relax}
\providecommand{\BIBforeignlanguage}[2]{{%
\expandafter\ifx\csname l@#1\endcsname\relax
\typeout{** WARNING: IEEEtran.bst: No hyphenation pattern has been}%
\typeout{** loaded for the language `#1'. Using the pattern for}%
\typeout{** the default language instead.}%
\else
\language=\csname l@#1\endcsname
\fi
#2}}
\providecommand{\BIBdecl}{\relax}
\BIBdecl

\bibitem{Dong&DiScenna&Guerra.Tupodb.2019}
X.~Dong, M.~DiScenna, and E.~Guerra, ``Transit user perceptions of driverless
  buses,'' \emph{Transportation}, vol.~46, no.~1, pp. 35--50, 2019.

\bibitem{Montes&Salinas.AEPfABD.2017}
H.~Montes, C.~Salinas, R.~Fern{\'a}ndez, and M.~Armada, ``An experimental
  platform for autonomous bus development,'' \emph{Applied Sciences}, vol.~7,
  no.~11, p. 1131, 2017.

\bibitem{Millonig&Frohlich.WABMaMNBtGit4AsoPTPN.09232018}
A.~Millonig and P.~Fr{\"o}hlich, ``Where autonomous buses might and might not
  bridge the gaps in the 4 a's of public transport passenger needs,'' in
  \emph{Proceedings of the 10th International Conference on Automotive User
  Interfaces and Interactive Vehicular Applications}.\hskip 1em plus 0.5em
  minus 0.4em\relax New York, NY, USA: ACM, 09232018, pp. 291--297.

\bibitem{Dosovitskiy&Ros.CAOUDS.2017}
A.~Dosovitskiy, G.~Ros, F.~Codevilla, A.~Lopez, and V.~Koltun, ``Carla: An open
  urban driving simulator,'' 2017.

\bibitem{Dworak&Ciepiela.PoLoddlaboagpcdfCs.20198262019829}
D.~Dworak, F.~Ciepiela, J.~Derbisz, I.~Izzat, M.~Komorkiewicz, and M.~Wojcik,
  ``Performance of lidar object detection deep learning architectures based on
  artificially generated point cloud data from carla simulator,'' in \emph{2019
  24th International Conference on Methods and Models in Automation and
  Robotics (MMAR)}.\hskip 1em plus 0.5em minus 0.4em\relax IEEE, 2019/8/26 -
  2019/8/29, pp. 600--605.

\bibitem{Iclodean&Cordos&Varga.ASBfPTAR.2020}
C.~Iclodean, N.~Cordos, and B.~O. Varga, ``Autonomous shuttle bus for public
  transportation: A review,'' \emph{Energies}, vol.~13, no.~11, p. 2917, 2020.

\bibitem{Bosch&Becker.Cbaoams.2018}
P.~M. B{\"o}sch, F.~Becker, H.~Becker, and K.~W. Axhausen, ``Cost-based
  analysis of autonomous mobility services,'' \emph{Transport Policy}, vol.~64,
  pp. 76--91, 2018.

\bibitem{Leich&Bischoff.SastrbAss.2018}
G.~Leich and J.~Bischoff, ``Should autonomous shared taxis replace buses? a
  simulation study,'' \emph{Transportation Research Procedia}, vol.~41, pp.
  450--460, 2018.

\bibitem{Portouli&Karaseitanidis.PatamboirciaHc.2017}
E.~Portouli, G.~Karaseitanidis, P.~Lytrivis, A.~Amditis, O.~Raptis, and
  C.~Karaberi, ``Public attitudes towards autonomous mini buses operating in
  real conditions in a hellenic city,'' in \emph{2017 IEEE Intelligent Vehicles
  Symposium (IV)}, 2017, pp. 571--576.

\bibitem{Lam&Katupitiya.Macoapoab.20136232013626}
S.~Lam and J.~Katupitiya, ``Modeling and control of a platoon of autonomous
  buses,'' in \emph{2013 IEEE Intelligent Vehicles Symposium (IV)}.\hskip 1em
  plus 0.5em minus 0.4em\relax IEEE, 2013/6/23 - 2013/6/26, pp. 958--963.

\bibitem{Gruyer&Magnier.PipamCsfada.2017}
D.~Gruyer, V.~Magnier, K.~Hamdi, L.~Claussmann, O.~Orfila, and
  A.~Rakotonirainy, ``Perception, information processing and modeling: Critical
  stages for autonomous driving applications,'' \emph{Annual Reviews in
  Control}, vol.~44, no.~3, pp. 323--341, 2017.

\bibitem{Chen&Jian.Adccasu.2019}
S.~Chen, Z.~Jian, Y.~Huang, Y.~Chen, Z.~Zhou, and N.~Zheng, ``Autonomous
  driving: cognitive construction and situation understanding,'' \emph{Science
  China Information Sciences}, vol.~62, no.~8, p. 661, 2019.

\bibitem{LiW.PanC.W.ZhangR.RenJ.P.MaY.X.FangJ.YanF.L.GengQ.C.HuangX.Y.Gon.2AAadsudda42.2019}
W.~Li, C.~W. Pan, R.~Zhang, J.~P. Ren, Y.~X. Ma, J.~Fang, F.~L. Yan, Q.~C.
  Geng, X.~Y. Huang, H.~J. Gong, and W.~W. Xu, ``Aads: Augmented autonomous
  driving simulation using data-driven algorithms. , 4(28).'' \emph{Science
  robotics}, 2019.

\bibitem{Azad&Hoseinzadeh.FABALRaFRD.2019}
M.~Azad, N.~Hoseinzadeh, C.~Brakewood, C.~R. Cherry, and L.~D. Han, ``Fully
  autonomous buses: A literature review and future research directions,''
  \emph{Journal of Advanced Transportation}, vol. 2019, no.~3, pp. 1--16, 2019.

\bibitem{Ainsalu&Arffman.SotAoAB.2018}
J.~Ainsalu, V.~Arffman, M.~Bellone, M.~Ellner, T.~Haapam{\"a}ki, N.~Haavisto,
  E.~Josefson, A.~Ismailogullari, E.~Pilli-Sihvola, O.~Madland, R.~Madzulaitis,
  J.~Muur, S.~M{\"a}kinen, V.~Nousiainen, E.~Rutanen, S.~Sahala,
  B.~Sch{\o}nfeldt, P.~M. Smolnicki, R.-M. Soe, J.~S{\"a}{\"a}ski,
  M.~Szyma{\'n}ska, I.~Vaskinn, and M.~{\AA}man, \emph{State of the Art of
  Automated Buses}, 2018.

\end{thebibliography}
\bibliographystyle{IEEEtran}

\end{document}